%% file: main.tex
\documentclass[10pt,twocolumn,letterpaper]{article}

\usepackage{cvpr}              % To produce the CAMERA-READY version
\usepackage{graphicx}
\usepackage{amsmath}
\usepackage{amssymb}
\usepackage{booktabs}
\usepackage{amsmath}
\usepackage{amssymb}
\usepackage{siunitx}
\usepackage{cite}
\usepackage{ctable}
\usepackage{hhline}
\usepackage{booktabs}
\usepackage{subcaption}
\usepackage{csquotes}
\usepackage[super]{nth}
\usepackage{multirow}
\usepackage{makecell}
\usepackage{arydshln}
\usepackage{pifont}
\usepackage{xcolor}
\usepackage{paralist}

\usepackage[pagebackref,breaklinks,colorlinks]{hyperref}
\newcommand{\quotes}[1]{``#1''}
\DeclareMathAlphabet{\numberbb}{U}{BOONDOX-ds}{m}{n}
\newcolumntype{L}[1]{>{\raggedright\let\newline\\\arraybackslash\hspace{0pt}}m{#1}}
\newcolumntype{C}[1]{>{\centering\let\newline\\\arraybackslash\hspace{0pt}}m{#1}}
\newcolumntype{R}[1]{>{\raggedleft\let\newline\\\arraybackslash\hspace{0pt}}m{#1}}

\usepackage[capitalize]{cleveref}
\crefname{section}{Sec.}{Secs.}
\Crefname{section}{Section}{Sections}
\Crefname{table}{Table}{Tables}
\crefname{table}{Tab.}{Tabs.}

\def\confName{CVPR}
\def\confYear{2022}

\begin{document}
\pagestyle{plain}

\title{NeuralAnnot: Neural Annotator for 3D Human Mesh Training Sets}

\author{
Gyeongsik Moon$^{1}$\hspace{1.0cm} Hongsuk Choi$^{1}$\hspace{1.0cm} Kyoung Mu Lee$^{1,2}$\\
\\
$^{1}$Dept. of ECE \& ASRI, $^{2}$IPAI, Seoul National University, Korea \hspace{1.0cm}
\\
{\small \texttt {\{mks0601,redarknight,kyoungmu\}@snu.ac.kr}}
}

\maketitle

%%%%%%%%% ABSTRACT
\begin{abstract}
Most 3D human mesh regressors are fully supervised with 3D pseudo-GT human model parameters and weakly supervised with GT 2D/3D joint coordinates as the 3D pseudo-GTs bring great performance gain.
The 3D pseudo-GTs are obtained by annotators, systems that iteratively fit 3D human model parameters to GT 2D/3D joint coordinates of training sets in the pre-processing stage of the regressors.
The fitted 3D parameters at the last fitting iteration become the 3D pseudo-GTs, used to fully supervise the regressors.
Optimization-based annotators, such as SMPLify-X, have been widely used to obtain the 3D pseudo-GTs.
However, they often produce wrong 3D pseudo-GTs as they fit the 3D parameters to GT of each sample independently.
To overcome the limitation, we present NeuralAnnot, a neural network-based annotator.
The main idea of NeuralAnnot is to employ a neural network-based regressor and dedicate it for the annotation.
Assuming no 3D pseudo-GTs are available, NeuralAnnot is weakly supervised with GT 2D/3D joint coordinates of training sets.
The testing results on the same training sets become 3D pseudo-GTs, used to fully supervise the regressors.
We show that 3D pseudo-GTs of NeuralAnnot are highly beneficial to train the regressors.
We made our 3D pseudo-GTs publicly available.
\end{abstract}

\input{src/introduction}
\input{src/related_works}

\input{src/neuralannot}

\input{src/body_hand_face}

\input{src/experiment}

\input{src/conclusion}

\noindent\textbf{Acknowledgements.}
This work was supported in part by IITP grant funded by the Korea government (MSIT) [No. 2021-0-01343, Artificial Intelligence Graduate School Program (Seoul National University)]

\input{src/suppl}

\clearpage

%%%%%%%%% REFERENCES
{\small
\bibliographystyle{ieee_fullname}
\bibliography{main}
}

\end{document}

%% file: src/introduction.tex
\section{Introduction}~\label{sec:intro}

3D human mesh estimation aims to localize human mesh vertices in the 3D space.
Two types of datasets have been used by most 3D human mesh regressors~\cite{kanazawa2018end,kolotouros2019learning,moon2020i2l,choutas2020monocular,rong2021frankmocap,moon2022hand4whole}.
The first dataset type is motion capture (MoCap) datasets~\cite{ionescu2014human3,mehta2017monocular,joo2015panoptic,yu2020humbi,moon2020interhand2}, captured with well-calibrated multiple cameras in highly restricted environment (\textit{e.g.}, MoCap studio).
Groundtruth (GT) 3D joint coordinates can be obtained owing to the special equipment; however, the restricted environment makes captured images have monotonous appearances.
The second dataset type is in-the-wild datasets~\cite{lin2014microsoft,jin2020whole}.
They have images with diverse appearances as they are captured in our daily life without the special equipment.
However, only GT 2D joint coordinates can be obtained by manual human labor, and 3D GTs are not obtainable because of depth and scale ambiguity.
The GT 2D/3D joint coordinates are used to weakly supervise the 3D human mesh regressors.

\begin{table}[t]
\small
\centering
\setlength\tabcolsep{1.0pt}
\def\arraystretch{1.1}
\begin{tabular}{C{2.8cm}|C{3.3cm}C{2.0cm}}
\specialrule{.1em}{.05em}{.05em}
Methods & Supervision targets & Where to test \\ \hline
\multirow{2}{*}{\textbf{\shortstack[c]{One-stage annotator\\(~\cite{pavlakos2019expressive},Ours)}}} & \multirow{2}{*}{\textbf{GT 2D/3D joint coords.}} & \multirow{2}{*}{\textbf{Training set}} \\ 
& & \\
\multirow{2}{*}{\shortstack[c]{Two-stage annotator\\~\cite{kolotouros2019learning,joo2020exemplar}}} & \multirow{2}{*}{\shortstack[c]{GT 2D/3D joint coords.\\+ initial 3D pseudo-GTs}} & \multirow{2}{*}{\textbf{Training set}} \\
& & \\
\multirow{2}{*}{Regressor~\cite{kanazawa2018end,kolotouros2019convolutional,moon2020i2l}} & \multirow{2}{*}{\shortstack[c]{GT 2D/3D joint coords.\\+ 3D pseudo-GTs}} & \multirow{2}{*}{Test set} \\
& & \\
\specialrule{.1em}{.05em}{.05em}
\end{tabular}
\vspace*{-3mm}
\caption{
Comparison of one-stage annotator (ours), two-stage annotator, and regressor.
}
\vspace*{-5mm}
\label{table:intro_compare}
\end{table}

In addition to the weak supervisions with GT 2D/3D joint coordinates, the regressors are fully supervised with 3D pseudo-GT human model parameters.
SMPL body model~\cite{loper2015smpl}, MANO hand model~\cite{romero2017embodied}, FLAME face model~\cite{li2017learning}, and SMPL-X~\cite{pavlakos2019expressive} whole-body model are widely used 3D human models.
Their parameters (\textit{i.e.}, 3D pseudo-GTs) include 3D joint rotations and latent codes of body shapes, hand shapes, or facial expressions.
\quotes{pseudo} represents that the parameters are obtained by being fit to GT 2D/3D joint coordinates; therefore, they are not true GTs.
Although the 3D pseudo-GTs can have fitting errors, they complement the GT 2D/3D joint coordinates in two ways.
First, 3D pseudo-GTs are full supervision targets, while GT 2D/3D joint coordinates are weak supervision targets, for the 3D human mesh regressors.
This is because 3D pseudo-GTs of the input human represent a single 3D human mesh without any ambiguity.
On the other hand, 2D/3D joint coordinates of the input human can represent multiple 3D human meshes.
For example, 3D joint coordinates cannot determine a single roll-axis rotation of each joint.
Furthermore, 2D joint coordinates suffer from severe depth ambiguity; therefore, multiple 3D meshes can be projected to the 2D joint coordinates.
%First, 3D mesh, obtainable by forwarding the 3D pseudo-GTs to 3D human model layers, give full supervision, while joint coordinates give weak supervision.
%This is because the 3D mesh can provide shape information (\textit{e.g.}, slim and fat) and 3D joint rotation information, not available in joint coordinates.
%For example, roll-axis rotations are not directly obtainable from 3D joint coordinates.
Second, 3D pseudo-GTs can provide 3D annotations for in-the-wild datasets.
As in-the-wild datasets contain only GT 2D joint coordinates without 3Ds, 3D pseudo-GTs of in-the-wild datasets can resolve the most major bottleneck of 3D human mesh estimation: lack of 3D data in the wild.
As the two points make 3D pseudo-GTs bring large performance improvements, most 3D human mesh estimation regressors~\cite{kanazawa2018end,kolotouros2019learning,moon2020i2l,choutas2020monocular,rong2021frankmocap} are fully supervised with 3D pseudo-GTs and weakly supervised with GT 2D/3D joint coordinates.

The 3D pseudo-GTs are obtained by annotators, systems that iteratively fit the 3D human model parameters to GT 2D/3D joint coordinates of training sets in the pre-processing stage of the regressors.
The fitted 3D parameters at the last fitting iteration become the 3D pseudo-GTs.
After fully supervised with the 3D pseudo-GTs and weakly supervised with GT 2D/3D joint coordinates, the regressors are tested on unseen samples in testing sets.
Table~\ref{table:intro_compare} shows comparison of one-stage annotator, two-stage annotator, and regressor.
The one-stage annotators, to which SMPLify-X~\cite{pavlakos2019expressive} belongs, output 3D pseudo-GTs without initial 3D pseudo-GTs.
SMPLify-X fits 3D human model parameters of T-pose to GT 2D joint coordinates of each sample.
Let us denote the annotation procedure of the one-stage annotator as $f_1$.
On the other hand, the two-stage annotators, to which SPIN~\cite{kolotouros2019learning} and EFT~\cite{joo2020exemplar} belong, first obtain initial 3D pseudo-GTs from the one-stage annotators in the pre-processing stage of the two-stage annotators. 
This is the first stage of the two-stage annotator, which is exactly same with the annotation procedure of the one-stage annotator: $f_1$.
Then, in the second stage, they are fully supervised with the initial 3D pseudo-GTs and weakly supervised with GT 2D/3D joint coordinates for the final 3D pseudo-GTs.
%In specific, SPIN predicts 3D human model parameters using their network and fits the predicted parameters to GT 2D joint coordinates of the input image using an optimization-based annotator.
%EFT fine-tunes the pre-trained SPIN network to GT 2D joint coordinates of the input image, and the outputs of the last fine-tuning iteration become a 3D pseudo-GT of the input image.
%Please note that SPIN and EFT can also be used as the regressors by testing them on unseen test sets.
%As a regressor, SPIN regresses 3D human mesh using their trained network without performing the fitting, and EFT fits the output of the SPIN network to predicted 2D joint coordinates without using GTs.
%In this paper, we focus only on their roles as annotators.

In the current literature, optimization-based annotators, such as SMPLify-X~\cite{pavlakos2019expressive}, are widely used.
However, they often produce wrong 3D pseudo-GTs as they fit the 3D parameters to each sample independently.
To overcome the limitation, we present NeuralAnnot, the first neural network-based one-stage annotator.
The main idea of NeuralAnnot is to employ a neural network-based regressor and dedicate it for the annotation.
Assuming no 3D pseudo-GTs are available, NeuralAnnot is weakly supervised with GT 2D/3D joint coordinates of training sets.
Then, it is tested on the same training set.
The testing results on the testing set become 3D pseudo-GTs, used to fully supervise the regressors.
We train and test NeuralAnnot in the pre-processing stage of the regressors.
NeuralAnnot's training and testing pipeline is clearly different from that of the regressors as the regressors are trained on training sets and tested on unseen test sets.
In contrast, we train and test NeuralAnnot on the same training set.
As it is tested on seen samples of the training set, there is no concern on the generalization performance on unseen data.
%NeuralAnnot shares a similar pipeline with recent state-of-the-art neural network-based regressors, which predict the 3D parameters from an image~\cite{kanazawa2018end,kolotouros2019learning,moon2020i2l}.
%However, NeuralAnnot, as a one-stage annotator, has three different points compared to the recent regressors.
%First, NeuralAnnot is trained and tested on the same training set, while the regressors are trained in training sets and tested on unseen test sets.
%Second, NeuralAnnot is trained and tested in the pre-processing stage of the regressors.
%After the pre-processing stage, the outputs of NeuralAnnot (\textit{i.e.}, 3D pseudo-GTs) are used to train the regressors.
%Third, NeuralAnnot is supervised with GT 2D/3D joint coordinates without 3D pseudo-GTs as its goal is to produce the 3D pseudo-GTs.
%On the other hand, the regressors are trained with GT 2D/3D joint coordinates in addition to the 3D pseudo-GTs.
%NeuralAnnot is the first attempt to dedicate a neural network to obtain the 3D pseudo-GTs.
%The data-driven learning of neural networks enables NeuralAnnot to extract deep semantic features, which provide highly useful cues to produce 3D pseudo-GTs.
Like SMPLify-X, our NeuralAnnot is a one-stage annotator, which does not require initial 3D pseudo-GTs and produces 3D pseudo-GTs only from GT 2D/3D joint coordinates.
%Different from SPIN and EFT that can be used as regressors, NeuralAnnot is designed only for the annotation; therefore, we do not test NeuralAnnot to unseen test sets.

In our experiments, we observed that simply employing a regressor for the annotation does not result in high-quality 3D pseudo-GTs.
In particular, in-the-wild datasets only provide GT 2D joint coordinates without 3Ds; therefore, overcoming the depth ambiguity is one of the major bottlenecks for the annotation of in-the-wild datasets.
We investigate the best combinations of inputs and outputs for the annotation of in-the-wild datasets and MoCap datasets.

For a fair comparison, we pick SMPLify-X~\cite{pavlakos2019expressive} for our main comparison target as it is a one-stage annotator like our NeuralAnnot and the most widely used one.
We additionally compare ours with SPIN and EFT in the experimental section after changing NeuralAnnot to a two-stage annotator.
All evaluation metrics measure how the produced 3D pseudo-GTs of training sets are accurate and beneficial.
We show that our NeuralAnnot produces 3D pseudo-GTs with much higher quality than SMPLify-X.
We use NeuralAnnot to obtain 3D pseudo-GTs of the human body, hands, face, and whole body.
All our 3D pseudo-GTs will be publicly available.

Our contributions can be summarized as follows.
\begin{itemize}
\item We present NeuralAnnot, the first neural network-based one-stage annotator.
NeuralAnnot is trained and tested on the same training to produce 3D pseudo-GTs, used to fully supervise the regressors.
\item Our NeuralAnnot produces far more accurate and beneficial 3D pseudo-GTs than previous optimization-based annotators owing to the data-driven learning of neural networks.
\item The newly obtained 3D pseudo-GTs will be publicly released.
We believe the new 3D pseudo-GTs will be highly beneficial to the community.
\end{itemize}

%% file: src/related_works.tex
\section{Related works}

\noindent\textbf{Optimization-based 3D pseudo-GT annotators.}
The optimization-based annotators fit 3D human model parameters for target 2D/3D joint coordinates, 3D point clouds, or 3D scans, which become 3D pseudo-GTs.
They fit 3D human model parameters to each sample without considering other samples.
SMPLify~\cite{bogo2016keep} fits SMPL~\cite{loper2015smpl} to a given 2D joint coordinates by minimizing a 2D loss and several prior terms.
Joo~\etal~\cite{joo2018total} constructed the Total Capture dataset by fitting a whole-body 3D human model (\textit{i.e.}, Adam and Frank) to estimated 3D joint coordinates and 3D point clouds.
SMPLify-X~\cite{pavlakos2019expressive} extended SMPLify for the 3D human whole-body model optimization by proposing a new 3D human whole-body model, SMPL-X.
Von~\etal~\cite{von2018recovering} constructed 3DPW dataset by fitting SMPL parameters to detected 2D joint coordinates and IMU sensor values.
Zimmermann~\etal~\cite{Freihand2019} constructed FreiHAND dataset by fitting MANO parameters to multi-view 2D hand joint coordinates, 3D hand joint coordinates, and segmentations.
Kulon~\etal~\cite{kulon2020weakly} constructed Youtube3DHand dataset by fitting MANO parameters to detected 2D hand joint coordinates.
Patel~\etal~\cite{Patel:CVPR:2021} constructed AGORA dataset by fitting SMPL and SMPL-X parameters to 3D scans.
Among them, SMPLify-X~\cite{pavlakos2019expressive} is the most widely used annotator to obtain 3D pseudo-GTs.

\noindent\textbf{Neural network-based 3D pseudo-GT annotators.}
SPIN~\cite{kolotouros2019learning} and EFT~\cite{joo2020exemplar} are recently introduced neural network-based annotators.
SPIN first predicts SMPL parameters using a neural network and runs an optimization-based annotator~\cite{bogo2016keep}, which fits the predicted SMPL parameters to GT 2D joint coordinates.
Their final 3D pseudo-GT of each sample is obtained by selecting one with smaller SMPLify loss~\cite{bogo2016keep} between the outputs of their optimization-based annotator and prepared initial 3D pseudo-GTs.
The initial 3D pseudo-GTs are prepared before training their network by running SMPLify on GT 2D joint coordinates.
EFT~\cite{joo2020exemplar} fine-tunes the pre-trained network of SPIN to the GT 2D joint coordinates of each sample, and the outputs of the last fine-tuning iteration become 3D pseudo-GT of the sample.

One critical difference between our NeuralAnnot and SPIN/EFT is that ours is a one-stage annotator, while SPIN/EFT are two-stage annotators.
The one-stage annotators annotate 3D pseudo-GTs by being weakly supervised with GT 2D/3D joint coordinates without requiring initial 3D pseudo-GTs.
On the other hand, the two-stage annotators obtain initial 3D pseudo-GTs using a one-stage annotator in the pre-processing stage of the two-stage annotators.
SPIN and EFT use optimization-based annotators~\cite{bogo2016keep,pavlakos2019expressive} as an one-stage annotator to obtain the initial 3D pseudo-GTs.
Then, they are fully supervised with the initial 3D pseudo-GTs and weakly supervised with GT 2D/3D joint coordinates in the second stage to obtain the final 3D pseudo-GTs.
Due to the difference between the one-stage and two-stage annotators, SPIN and EFT are not our direct comparison targets.
Instead, our NeuralAnnot can improve their annotation pipeline by replacing the optimization-based annotators in their pre-processing stage.
Nevertheless, to show the clear benefit of NeuralAnnot, we compare ours with SPIN and EFT in the experimental section after changing NeuralAnnot to a two-stage annotator.
The comparison shows that the two-stage NeuralAnnot produces much more beneficial 3D pseudo-GTs than SPIN and EFT.

%% file: src/neuralannot.tex
\section{NeuralAnnot}

\subsection{Network architecture}
NeuralAnnot predicts 3D human model parameters $\Theta$ from a single image $\mathbf{I}$.
We design NeuralAnnot as a simple combination of ResNet-50~\cite{he2016deep} and a fully-connected layer $f$, which extracts image feature vector $\mathbf{f}$ and regresses a set of 3D human model parameters $\Theta$, respectively.
The 3D human model parameter set $\Theta$ is different for each human model, which will be described in Section~\ref{sec:body_hand_face}.
As our main focus is not proposing a new network architecture, we use the most widely used network architecture~\cite{kanazawa2018end,kolotouros2019learning}, a combination of ResNet and fully-connected layers.
%More advanced network architectures can be plugged into our NeuralAnnot framework for better 3D pseudo-GTs.

\begin{figure}[t]
\begin{center}
\includegraphics[width=\linewidth]{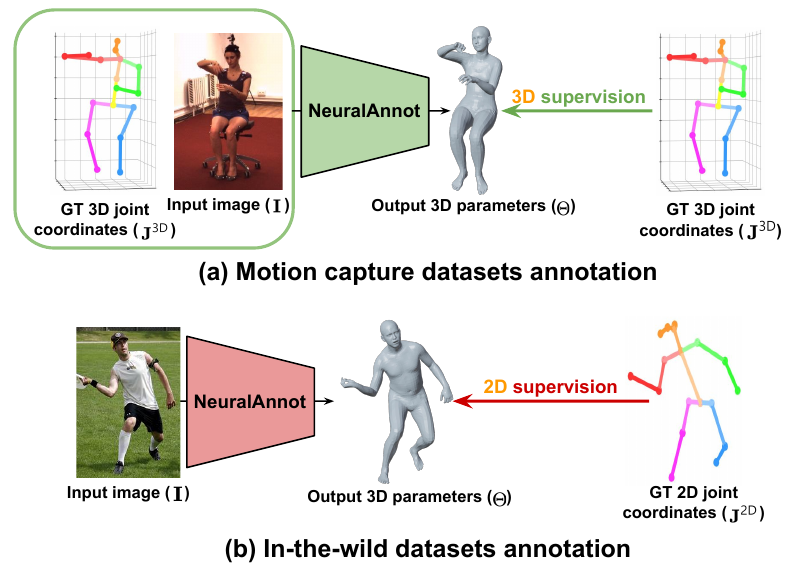}
\end{center}
\vspace*{-5mm}
   \caption{The overall pipeline of NeuralAnnot.
   (a) For the MoCap dataset annotation, it takes a pair of (image, GT 3D joint coordinates) and is supervised with GT 3D joint coordinates.
   (b) For the in-the-wild dataset annotation, it takes an image and is supervised with GT 2D joint coordinates.}
\vspace*{-5mm}   
\label{fig:overall_pipeline}
\end{figure}

\subsection{MoCap datasets annotation}
The MoCap datasets provide images with monotonous appearances and GT 3D joint coordinates $\mathbf{J}^\text{3D}$.
NeuralAnnot obtains 3D pseudo-GTs of MoCap datasets from scratch without initial 3D pseudo-GTs, which we call the one-stage annotation.
To this end, we provide the image feauture vector $\mathbf{f}$ and GT 3D joint coordinates $\mathbf{J}^\text{3D}$ to our network, as shown in Figure~\ref{fig:overall_pipeline} (a).
Please note that 3D pseudo-GTs consist of 3D joint \emph{rotations} and latent codes of 3D human models, which are not directly obtainable from 3D joint \emph{coordinates}.
The image feature vector $\mathbf{f}$ provides human articulation and shape information, while GT 3D joint coordinates $\mathbf{J}^\text{3D}$ additionally provide depth and real scale information, which 2D image features lack.
To this end, the GT 3D joint coordinates $\mathbf{J}^\text{3D}$ are converted to a 512-dimensional feature by two fully connected layers and concatenated with global average pooled ResNet output feature $\mathbf{f}$.
The concatenated feature is fed to a fully connected layer $f$ for the 3D human model parameter regression.
The 3D rotations of joints in the predicted 3D human model parameters are initially predicted in a 6D rotational representation of Zhou~\etal~\cite{zhou2019continuity} and converted to a 3D axis-angle rotation.
The loss function is defined as follows:
\begin{equation}
L_\text{mocap} = \| \hat{\mathbf{J}}^\text{3D} - \mathbf{J}^\text{3D} \|_1 + \sum_{\theta \in \Theta}{\lambda_{\text{mocap},\theta}\hat{\theta}^2},  
\vspace*{-1mm}
\end{equation}~\label{eq:mocap_loss}
where the hat mark indicates a predicted output.
The first term is a weak supervision from GT 3D joint coordinates, and the second term is a regularizer.
$\hat{\mathbf{J}}^\text{3D}$ is obtained by multiplying a joint regression matrix of a human model to a 3D mesh, where the 3D mesh is obtained by forwarding the 3D human model parameters $\Theta$ to a 3D human model layer.
$\lambda_{\text{mocap},\theta}$ denotes $L2$ norm regularizer weight of each 3D human parameter $\theta$.
The $L2$ norm regularizer prevents implausible 3D human mesh, widely used in previous works~\cite{bogo2016keep,pavlakos2019expressive}.
The network's testing outputs $\Theta$ on the training set become the 3D pseudo-GTs.

\subsection{In-the-wild datasets annotation}
The in-the-wild datasets provide images with diverse appearances and GT 2D joint coordinates $\mathbf{J}^\text{2D}$.
NeuralAnnot obtains 3D pseudo-GTs of in-the-wild datasets from scratch without initial 3D pseudo-GTs, which we call the one-stage annotation.
To this end, a fully-connected layer $f$ takes the image feature vector $\mathbf{f}$ and outputs 3D human model parameters $\Theta$, as shown in Figure~\ref{fig:overall_pipeline} (b).
Different from the MoCap datasets annotation scenario, NeuralAnnot takes a single image as an input without additional joint coordinates, of which the reason is reported in the experimental section.
As the in-the-wild datasets provide only GT 2D joint coordinates without 3D data, only 2D supervision is allowed when training NeuralAnnot's network, which can lead to implausible 3D human mesh due to the depth ambiguity.
To prevent NeuralAnnot's network from generating implausible 3D human meshes, we design it to predict a low-dimensional embedding of the 3D rotations (\textit{i.e.}, a latent code of VPoser~\cite{pavlakos2019expressive} for the body part and PCA coefficients for the hand part~\cite{romero2017embodied}), following SMPLify-X~\cite{pavlakos2019expressive}, unlike directly predicting 3D joint rotations when trained on MoCap datasets.
The low-dimensional embedding can effectively limit the output space to plausible 3D human articulation space, thus can prevent implausible 3D human mesh.
The loss function is defined as follows:
\begin{equation}
L_\text{wild} = \| \hat{\mathbf{J}}^\text{2D} - \mathbf{J}^\text{2D} \|_1 + \sum_{\theta \in \Theta}{\lambda_{\text{wild},\theta}\hat{\theta}^2},  
\vspace*{-1mm}
\end{equation}
where the hat mark indicates a predicted output.
The first term is a weak supervision from GT 2D joint coordinates, and the second term is a regularizer.
$\hat{\mathbf{J}}^\text{2D}$ is obtained by projecting 3D joint coordinates $\hat{\mathbf{J}}^\text{3D}$ to the image plane with predicted camera parameters.
$\lambda_{\text{wild},\theta}$ denotes $L2$ norm regularizer weight of each 3D human model parameter $\theta$.
Like when training our network on MoCap datasets, we used $L2$ norm regularizer to prevent implausible 3D human mesh.
The network's testing outputs $\Theta$ on the training set become the 3D pseudo-GTs.

%% file: src/body_hand_face.tex
\vspace*{-3mm}

\section{3D pseudo-GTs of body, hands, and face}~\label{sec:body_hand_face}
We use NeuralAnnot to obtain the 3D body, hands, face, and whole-body pseudo-GTs.
We provide how to obtain those 3D pseudo-GTs below.

\noindent\textbf{Body.}
We use SMPL~\cite{loper2015smpl} and SMPL-X~\cite{pavlakos2019expressive} as 3D body models.
To obtain SMPL 3D pseudo-GTs, our network predicts 3D body global rotation $\theta^g_b \in \mathbb{R}^3$, 3D body rotations $\theta_b \in \mathbb{R}^{21 \times 3}$, shape parameter $\beta_b \in \mathbb{R}^{10}$, and camera parameter $k_b \in \mathbb{R}^3$.
Our network for SMPL-X 3D pseudo-GTs predicts the same outputs; however, a joint set for 3D body rotation $\theta_b$ changes to that of SMPL-X.
Other SMPL-X parameters from hands and face are set to zero.
We set $\Theta$ to $\{\theta_b, \beta_b\}$ and $\{z_b, \beta_b\}$ when training NeuralAnnot's network for the MoCap datasets annotation and in-the-wild datasets annotation, respectively.
$z_b$ denotes the latent code of the VPoser~\cite{pavlakos2019expressive}.

\noindent\textbf{Hands.}
We use MANO~\cite{romero2017embodied} as a 3D hand model.
To obtain MANO 3D pseudo-GTs, our network predicts 3D hand global rotation $\theta^g_h \in \mathbb{R}^3$, 3D hand rotations $\theta_h \in \mathbb{R}^{15 \times 3}$, shape parameter $\beta_h \in \mathbb{R}^{10}$, and camera parameter $k_h \in \mathbb{R}^3$.
All hand images are flipped to the right hands, and we flip back the estimated pseudo-GTs of the left hands.
We set $\Theta$ to $\{\theta_h, \beta_h\}$ and $\{z_h, \beta_h\}$ when training NeuralAnnot's network for the MoCap datasets annotation and in-the-wild datasets annotation, respectively.
$z_h$ denotes the 3D hand pose PCA coefficients, defined in MANO.

\noindent\textbf{Face.}
We use FLAME~\cite{li2017learning} as a 3D face model.
To obtain FLAME 3D pseudo-GTs, our network predicts 3D face global rotation $\theta^g_f \in \mathbb{R}^3$, 3D jaw rotation $\theta_f \in \mathbb{R}^3$, shape parameter $\beta_f \in \mathbb{R}^{10}$, and expression code $\psi \in \mathbb{R}^{10}$.
We set $\Theta$ to $\{\theta_f, \beta_f, \psi\}$.

\noindent\textbf{Integration to whole body.}
After obtaining the body, hands, and face 3D pseudo-GTs, we get the final whole-body 3D pseudo-GTs by forwarding $\{\theta^g_b, \theta_b, \beta_b, \theta^g_{rh}, \theta_{rh}, \theta^g_{lh}, \theta_{lh}, \theta_f, \psi\}$ to SMPL-X, where $*_{rh}$ and $*_{lh}$ denote $*$ is from right and left hand, respectively.
$\theta^g_b$, $\theta_b$, and $\beta_b$ are from SMPL-X 3D pseudo-GTs.
The 3D hand rotations $\theta_h$ of MANO and 3D jaw rotation $\theta_f$ and facial expression code $\psi$ of FLAME are compatible with those of SMPL-X; thus, we use them for the final 3D pseudo-GTs.
As $\theta^g_h$ are not 3D local rotations in the human body kinematic chain, we change their 3D global rotations to 3D local rotations by multiplying the inverse of 3D elbow global rotations.

%% file: src/experiment.tex
\section{Experiment}

\subsection{Implementation details}

PyTorch~\cite{paszke2017automatic} is used for implementation. 
The backbone part is initialized with the publicly released ResNet50~\cite{he2016deep}, pre-trained on ImageNet~\cite{russakovsky2015imagenet}. 
The weights are updated by the Adam optimizer~\cite{kingma2014adam} with a mini-batch size of 192.
All input images are cropped using GT box and resized to 256$\times$256.
%No data augmentation is performed as our goal is to overfit NeuralAnnot to the training sets.
The initial learning rate is set to $10^{-4}$ and reduced by a factor of 10 when it converges.

\begin{figure}[t]
\begin{center}
\includegraphics[width=\linewidth]{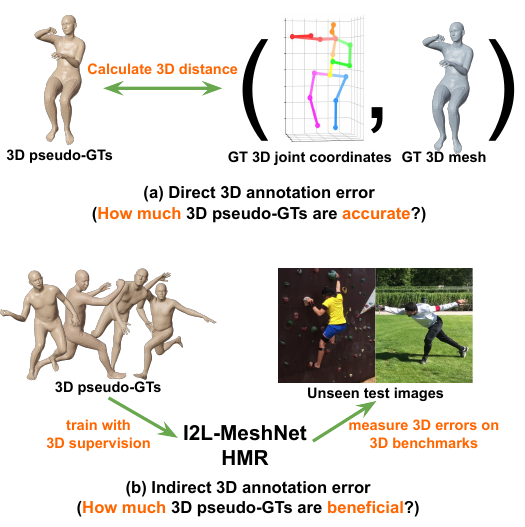}
\end{center}
\vspace*{-5mm}
   \caption{Illustrations of our metrics to evaluate annotators.
   (a) Direct 3D annotation error measures MPJPE and MPVPE, only applicable to datasets that provide 3D GTs.
   (b) Indirect 3D annotation error measures 3D errors of regressors, such as I2L-MeshNet~\cite{moon2020i2l} and HMR~\cite{kanazawa2018end}, on 3D benchmarks after training them on the obtained 3D pseudo-GTs of MSCOCO.
   This is used for in-the-wild datasets as the absence of 3D GTs makes it impossible to calculate the direct 3D annotation error.
   }
\vspace*{-3mm}
\label{fig:evaluation_metric}
\end{figure}

\subsection{Datasets for the annotation}\noindent
We use various datasets for the annotation as described below.
Please note that in addition to the datasets below, our NeuralAnnot can be used for any datasets that provide GT 2D/3D joint coordinates.

\noindent\textbf{MoCap datasets.}
We annotate 3D body pseudo-GTs of Human3.6M~\cite{ionescu2014human3}, MPI-INF-3DHP~\cite{mehta2017monocular}, and 3DPW~\cite{von2018recovering}.
Also, 3D hand pseudo-GTs of InterHand2.6M~\cite{moon2020interhand2} and FreiHAND~\cite{Freihand2019} are annotated.
Among them, only 3DPW and FreiHAND have GT 3D meshes, and all of them have only GT 3D joint coordinates without meshes.

\noindent\textbf{In-the-wild datasets.}
We annotate 3D body, hands, face, and whole-body pseudo-GTs of MSCOCO~\cite{lin2014microsoft,jin2020whole}.
It only provides GT 2D joint coordinates without 3D ones.

\subsection{Evaluation metrics}~\label{sec:metric}

\noindent\textbf{Direct 3D annotation error.}
Figure~\ref{fig:evaluation_metric} (a) shows how we calculate the direct 3D annotation error.
The direct 3D annotation error is measured using mean per joint position error (MPJPE) and mean per-vertex position error (MPVPE), which are the average 3D joint distance (mm) and 3D mesh vertex distance (mm) between GT and ones from 3D pseudo-GTs after aligning a root joint position.
This metric is only applicable to datasets that provide GT 3D joint coordinates or GT 3D meshes.

\begin{figure*}[t]
\begin{center}
\includegraphics[width=\linewidth]{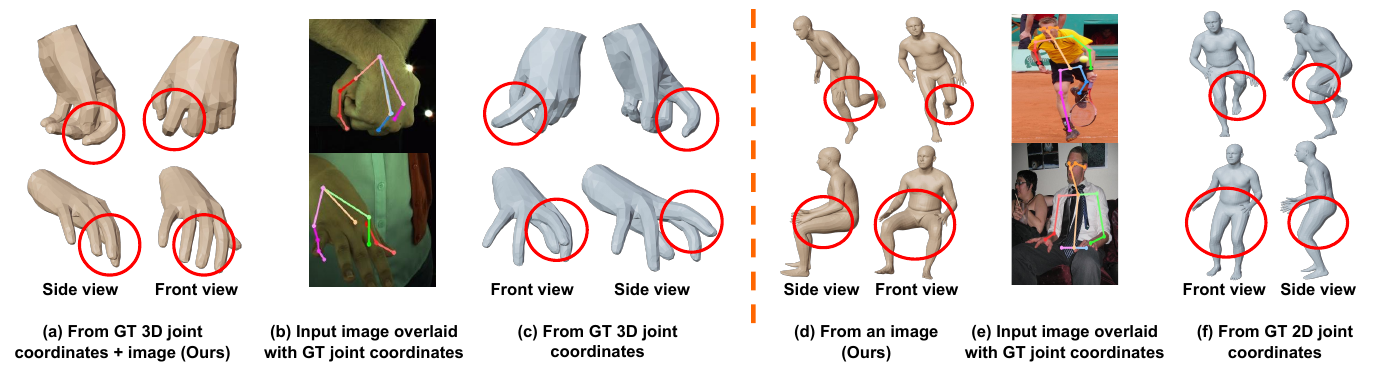}
\end{center}
\vspace*{-5mm}
   \caption{Qualitative comparison of 3D hand and body pseudo-GTs from NeuralAnnots that take various inputs on InterHand2.6M (left, a-c) and MSCOCO (right, d-f).
   The top row of results on InterHand2.6M is from a left hand, occluded by the right hand.
   GT 3D joint coordinates of the samples from InterHand2.6M are partially missing due to the capture failure.
   GT 2D joint coordinates of the bottom sample from MSCOCO are partially missing due to the truncation.
    }
\label{fig:neuralannot_input}
\vspace*{-4mm}
\end{figure*}

\noindent\textbf{Indirect 3D annotation error.}
Figure~\ref{fig:evaluation_metric} (b) shows how we designed the indirect 3D annotation error.
The indirect 3D annotation error represents how much the 3D pseudo-GTs are beneficial for the training.
The absence of the 3D GTs of the in-the-wild datasets makes calculating the direct 3D annotation error impossible.
Alternatively, we use 3D test errors of regressors after training them on 3D pseudo-GTs of MSCOCO as an indirect 3D annotation error.
The indirect 3D annotation error would decrease if the 3D pseudo-GTS from annotators become more beneficial so that the trained regressors work better.
We use I2L-MeshMet~\cite{moon2020i2l} and HMR~\cite{kanazawa2018end} as regressors for the indirect error calculation.
I2L-MeshNet and HMR are based on model-free and model-based approaches, which predict 3D vertex coordinates and 3D human model parameters, respectively.
We chose such very different regressors to minimize dependency on the type of the regressors so that the indirect error could be less stochastic.
For the indirect error of the body part, we use PA MPJPE (mm) and PA MPVPE (mm), which further align rotation and scale in addition to the translation.
For the hand part, area under curve (AUC) of a 3D mesh is used.
Finally, for the face part, a 3D surface error, an average distance between GT 3D scan and the closest 3D mesh vertex is used.

\subsection{Ablation study}
\noindent\textbf{Inputs of NeuralAnnot.}
For the MoCap datasets annotation, our network takes a pair of (image, GT 3D joint coordinates) as an input.
The Figure~\ref{fig:neuralannot_input} left (a-c) and first, second, and third rows of Table~\ref{table:neuralannot_input} validate our input choice for the MoCap datasets annotation.
The figure shows that taking an image as input is greatly helpful to recover partially missing GT 3D joint coordinates, which arises from the failure of the marker-less data capture.
The table shows that additional GT 3D joint coordinates reduce the direct 3D error by providing depth and scale information to the network.

For the in-the-wild datasets annotation, our network takes an image as an input.
The Figure~\ref{fig:neuralannot_input} right (d-f) and the fourth, fifth, and sixth rows of Table~\ref{table:neuralannot_input} validate our input choice for the in-the-wild datasets annotation.
The figure shows that taking an image suffers from less depth ambiguity as image appearances can provide depth information, which lacks in 2D joint coordinates.
For example, in the first row of the figure, the shadow and small size of the left knee tell us that the leg is behind the right leg.
In addition, the image input can provide contextual information, helpful when GT 2D joint coordinates are missing or truncated.
The table shows that 3D pseudo-GTs from the image input achieve lower indirect 3D annotation error.
We observed that using both an image and GT 2D joint coordinates as an input achieves similar indirect 3D error compared with a model that takes an image as an input, while raising computational costs.
We think the reason for the similar indirect 3D error is that the GT 2D joint coordinates do not offer additional depth information as GT 3D joint coordinates did in the MoCap datasets.
To achieve the best quality of 3D pseudo-GTs with smaller computational costs, we used only an image as an input of the network without GT 2D joint coordinates.
For the comparison, we provided GT 2D joint coordinates as a 2D Gaussian heatmap representation to our network following Moon~\etal~\cite{moon2019posefix}.

\begin{table}[t]
\small
\centering
\setlength\tabcolsep{1.0pt}
\def\arraystretch{1.1}
\begin{tabular}{L{5.7cm}|C{2.3cm}}
\specialrule{.1em}{.05em}{.05em}
 Inputs of NeuralAnnot & Annotation err. \\ \hline
  \textbf{* MoCap datasets annotation} & \textbf{Direct 3D err.} \\
 Image   & 7.0 \\
 GT 3D joint coordinates & 8.1 \\
 \textbf{GT 3D joint coordinates + Image (Ours)}  & \textbf{5.8} \\ \hline
 \textbf{* In-the-wild datasets annotation} & \textbf{Indirect 3D err.} \\
 \textbf{Image (Ours)} & \textbf{69.6} \\
 GT 2D joint coordinates   & 72.7 \\
 GT 2D joint coordinates + Image & \textbf{69.6} \\
 \specialrule{.1em}{.05em}{.05em}
\end{tabular}
\vspace*{-3mm}
\caption{
The annotation error comparison of NeuralAnnots that take different inputs.
The direct 3D errors (MPJPE) are computed on InterHand2.6M, and the indirect 3D errors (PA MPJPE) are computed on 3DPW after training I2L-MeshNet on 3D pseudo-GTs of MSCOCO.
}
\vspace*{-4mm}
\label{table:neuralannot_input}
\end{table}

\noindent\textbf{Outputs of NeuralAnnot.}
Table~\ref{table:neuralannot_output} validates our design for NeuralAnnot's output type.
For the MoCap datasets annotation, our network outputs 3D joint rotations while it outputs the low-dimensional embedded poses (\textit{e.g.}, VPoser of SMPL/SMPL-X and PCA coefficients of MANO) for the in-the-wild datasets annotation.
The low-dimensional embedded poses can be helpful when GT carries insufficient 3D information (\textit{e.g.}, 2D joint coordinates) by restricting outputs to learned latent pose space.
For example, it can prevent many anatomically implausible 3D meshes that correspond to 2D joint coordinates.
On the other hand, the low-dimensional embedded poses can be harmful when GT carries enough 3D information (\textit{e.g.}, 3D joint coordinates) as the 3D information of GT is enough to prevent the implausible 3D meshes, while some 3D poses cannot be represented in the learned latent pose space.
The first and second rows of the table show that when annotating MoCap datasets, estimating 3D joint rotations achieves better results due to the availability of GT 3D joint coordinates.
The third and fourth rows of the table show that when annotating in-the-wild datasets, estimating the low-dimensional embedded pose produces better results as only GT 2D joint coordinates are available without 3D ones.
We designed the network to produce the best output type considering which type of supervision is available for the annotation of each dataset.

\begin{table}[t]
\small
\centering
\setlength\tabcolsep{1.0pt}
\def\arraystretch{1.1}
\begin{tabular}{L{5.7cm}|C{2.3cm}}
\specialrule{.1em}{.05em}{.05em}
 Outputs of NeuralAnnot & Annotation err. \\ \hline
  \textbf{* MoCap datasets annotation} & \textbf{Direct 3D err.} \\
  \textbf{3D joint rotations (Ours)}  & \textbf{5.8} \\
 Low-dimensional embedded pose   & 10.7 \\ \hline
 \textbf{* In-the-wild datasets annotation} & \textbf{Indirect 3D err.} \\
   3D joint rotations & 117.5 \\
  \textbf{Low-dimensional embedded pose (Ours)}  & \textbf{69.6} \\
 \specialrule{.1em}{.05em}{.05em}
\end{tabular}
\vspace*{-3mm}
\caption{
The annotation error comparison of NeuralAnnots that produce different outputs.
The direct 3D errors (MPJPE) are computed on InterHand2.6M, and the indirect 3D errors (PA MPJPE) are computed on 3DPW after training I2L-MeshNet on 3D pseudo-GTs of MSCOCO.
}
\vspace*{-4mm}
\label{table:neuralannot_output}
\end{table}

\subsection{Comparison with previous annotators}
As described in Section~\ref{sec:intro}, for the fair comparison, we pick SMPLify-X~\cite{pavlakos2019expressive} as our main comparison target as it is a one-stage annotator like our NeuralAnnot and the most widely used one.
We additionally compare ours with two-stage annotators, such as SPIN~\cite{kolotouros2019learning} and EFT~\cite{joo2020exemplar}, after changing NeuralAnnot to the two-stage annotator.

\noindent\textbf{MoCap datasets annotation using GT 3D joint coordinates.}
Table~\ref{table:compare_3ddb_3dpose} shows that NeuralAnnot achieves much lower direct 3D annotation error than SMPLify-X~\cite{pavlakos2019expressive} on various MoCap datasets when the 3D pseudo-GTs are fit to GT 3D joint coordinates like Figure~\ref{fig:overall_pipeline} (a).
GT 3D meshes are not used for the annotation but used only for evaluation purposes.
Moreover, NeuralAnnot can recover 3D pseudo-GTs from partially missing and noisy GT 3D joint coordinates by utilizing image features, as shown in Figure~\ref{fig:neuralannot_input} left (a-c), while SMPLify-X cannot as it only utilizes GT 3D joint coordinates without the image input.
For the comparison, we modified official released codes of SMPLify-X to optimize it to GT 3D joint coordinates as the original one only considers 2D joint coordinates during the optimization.
Like NeuralAnnot, SMPLify-X optimizes 3D joint rotations, not low-dimensional embedded poses, for the MoCap datasets annotation.
We describe its detailed modifications in the supplementary material.
MPVPEs are reported only for datasets that provide GT 3D meshes.

\begin{table}[t]
\small
\centering
\setlength\tabcolsep{1.0pt}
\def\arraystretch{1.1}
\begin{tabular}{L{2.3cm}|C{2.5cm}|C{3.0cm}}
\specialrule{.1em}{.05em}{.05em}
Datasets & SMPLify-X & \textbf{NeuralAnnot (Ours)} \\ \hline
\textbf{* Body} &  &  \\
Human3.6M & 13.1 / N/A & \textbf{8.5} / N/A \\
MPI-INF-3DHP & 17.9 / N/A & \textbf{12.2} / N/A \\ 
3DPW & 19.2 / 38.0 & \textbf{10.7} / \textbf{11.1} \\ \hline
\textbf{* Hands} & & \\
InterHand2.6M & 10.0 / N/A & \textbf{5.8} / N/A  \\
FreiHAND & 6.6 / 7.3 & \textbf{3.6} / \textbf{4.0} \\
\specialrule{.1em}{.05em}{.05em}
\end{tabular}
\vspace*{-3mm}
\caption{The direct 3D annotation error (MPJPE/MPVPE) comparison between SMPLify-X and our NeuralAnnot on various MoCap datasets. \textbf{The 3D pseudo-GTs are obtained by being fit to GT 3D joint coordinates.}}
%\vspace*{-3mm}
 \label{table:compare_3ddb_3dpose}
\end{table}

\noindent\textbf{MoCap datasets annotation using GT 2D joint coordinates.}
Table~\ref{table:compare_3ddb_2dpose} shows that NeuralAnnot achieves much lower direct 3D annotation error than SMPLify-X~\cite{pavlakos2019expressive} on various MoCap datasets when the 3D pseudo-GTs are fit to GT 2D joint coordinates.
The table is to simulate an in-the-wild dataset annotation scenario, in which only GT 2D joint coordinates are available, and provide direct 3D annotation errors.
We annotate 3DPW and FreiHAND datasets as they contain diverse outdoor images like in-the-wild images of MSCOCO compared to other MoCap datasets, such as Human3.6M and InterHand2.6M
To this end, we use the pipeline of Figure~\ref{fig:overall_pipeline} (b) for the MoCap dataset annotation.
SMPLify-X fits 3D human model parameters to GT 2D joint coordinates.
NeuralAnnot takes a single image from a MoCap dataset and supervises the network with GT 2D joint coordinates.
MPVPEs are reported only for datasets that provide GT 3D meshes.

\begin{table}[t]
\small
\centering
\setlength\tabcolsep{1.0pt}
\def\arraystretch{1.1}
\begin{tabular}{L{2.3cm}|C{2.5cm}|C{3.0cm}}
\specialrule{.1em}{.05em}{.05em}
Datasets & SMPLify-X & \textbf{NeuralAnnot (Ours)} \\ \hline
\textbf{* Body} &  &  \\
3DPW & 147.35 / 206.62 & \textbf{60.1} / \textbf{68.3} \\ \hline
\textbf{* Hands} & & \\
FreiHAND & 8.5 / 9.3 & \textbf{7.2} / \textbf{7.7} \\ 
\specialrule{.1em}{.05em}{.05em}
\end{tabular}
\vspace*{-3mm}
\caption{The direct 3D annotation error (MPJPE/MPVPE) comparison between SMPLify-X and our NeuralAnnot on various MoCap datasets. \textbf{The 3D pseudo-GTs are obtained by being fit to GT 2D joint coordinates.}}
%\vspace*{-3mm}
\label{table:compare_3ddb_2dpose}
\end{table}

\begin{figure*}[t]
\begin{center}
\includegraphics[width=\linewidth]{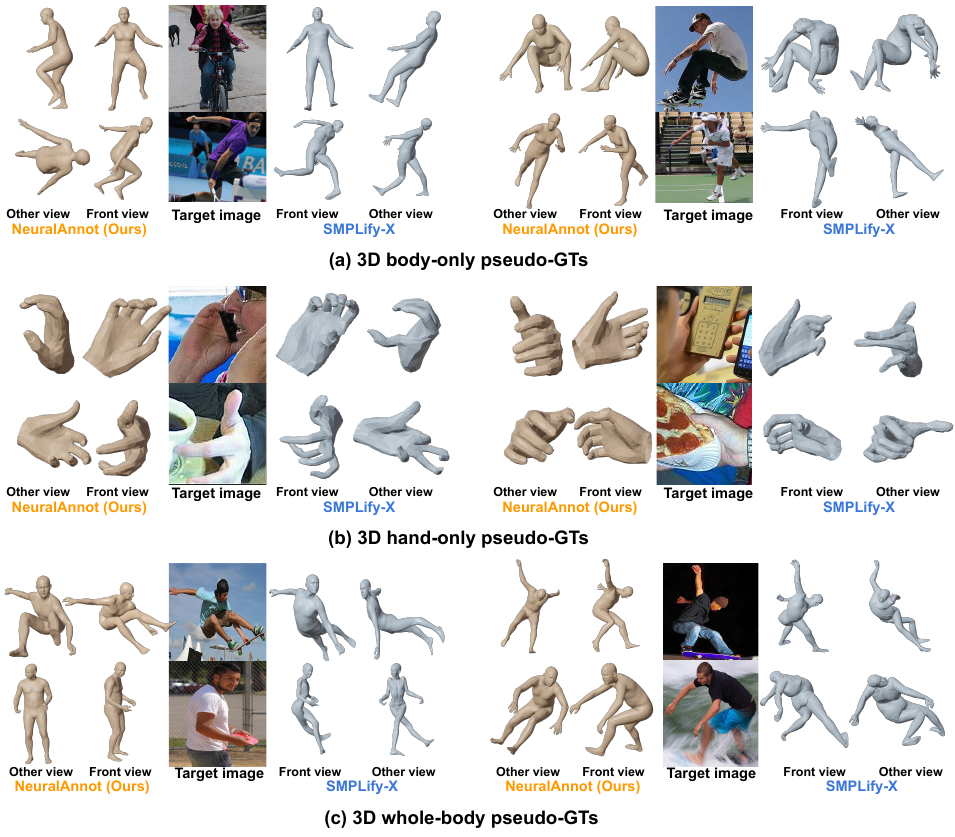}
\end{center}
\vspace*{-5mm}
   \caption{Qualitative comparison of 3D pseudo-GTs from NeuralAnnot and SMPLify-X.}
%\vspace*{-4mm}   
\label{fig:qualitative_comparison}
\end{figure*}

\noindent\textbf{In-the-wild datasets annotation using GT 2D joint coordinates.}
Table~\ref{table:compare_2ddb} shows that 3D pseudo-GTs from our NeuralAnnot achieve lower indirect 3D annotation error than SMPLify-X when the 3D pseudo-GTs are fit to GT 2D joint coordinates like Figure~\ref{fig:overall_pipeline} (b).
For the comparison, we ran SMPLify-X using their officially released codes.
It optimizes the low-dimensional embedded poses for the in-the-wild datasets annotation, like ours.
Figure~\ref{fig:qualitative_comparison} shows that SMPLify-X fails when the target image's pose has high depth ambiguity or there are truncations.
Unlike the body and hands, NeuralAnnot produces similar indirect 3D annotation error of 3D face pseudo-GTs compared with those of SMPLify-X.
We think the reason is that less complicated articulations of the face make SMPLify-X work well.

\begin{table}[t]
\small
\centering
\setlength\tabcolsep{1.0pt}
\def\arraystretch{1.1}
\begin{tabular}{L{3.3cm}|C{3.5cm}}
\specialrule{.1em}{.05em}{.05em}
Annotators  & Indirect 3D err.   \\ \hline
\textbf{* Body} & \textbf{PA MPVPE}$\downarrow$  \\
SMPLify-X & 80.7~\cite{moon2020i2l} / 81.4~\cite{kanazawa2018end} \\ 
\textbf{NeuralAnnot} \textbf{(Ours)} &  \textbf{71.6}~\cite{moon2020i2l} / \textbf{73.3}~\cite{kanazawa2018end} \\ \hline
\textbf{* Hands} & \textbf{AUC of 3D mesh}$\uparrow$ \\
SMPLify-X & 0.731~\cite{moon2020i2l} / 0.755~\cite{kanazawa2018end} \\ 
\textbf{NeuralAnnot (Ours)} & \textbf{0.760}~\cite{moon2020i2l} / \textbf{0.786}~\cite{kanazawa2018end} \\ \hline
\textbf{* Face} & \textbf{3D surface err.}$\downarrow$ \\
SMPLify-X & 2.05~\cite{moon2020i2l} / 2.06~\cite{kanazawa2018end} \\
\textbf{NeuralAnnot (Ours)}  & \textbf{2.03}~\cite{moon2020i2l} / \textbf{2.05}~\cite{kanazawa2018end}  \\
\specialrule{.1em}{.05em}{.05em}
\end{tabular}
\vspace*{-3mm}
\caption{The indirect 3D annotation errors comparison with one-stage annotator, SMPLify-X.
The indirect 3D annotations errors are computed on 3DPW, FreiHAND, and Stirling for the body, hands and face, respectively, after training I2L-MeshNet~\cite{moon2020i2l}/HMR~\cite{kanazawa2018end} on 3D pseudo-GTs of MSCOCO.}
%\vspace*{-3mm}
\label{table:compare_2ddb}
\end{table}

\begin{table}
\small
\centering
\setlength\tabcolsep{1.0pt}
\def\arraystretch{1.1}
\begin{tabular}{L{4.4cm}|C{3.0cm}}
\specialrule{.1em}{.05em}{.05em}
Annotators & Indirect 3D err.  \\ \hline
SPIN & 77.4~\cite{moon2020i2l} / 80.1~\cite{kanazawa2018end} \\ 
EFT & 71.2~\cite{moon2020i2l} / 72.1~\cite{kanazawa2018end} \\ 
\textbf{Two-stage NeuralAnnot (Ours)} & \textbf{64.6}~\cite{moon2020i2l} / \textbf{65.1}~\cite{kanazawa2018end} \\ \hline
\specialrule{.1em}{.05em}{.05em}
\end{tabular}
\vspace*{-3mm}
\caption{The indirect 3D annotation error (PA MPVPE) comparison with two-stage annotators.
The indirect 3D annotation errors are computed on 3DPW after training I2L-MeshNet~\cite{moon2020i2l}/HMR~\cite{kanazawa2018end} on 3D pseudo-GTs of MSCOCO.}
%\vspace*{-3mm}
\label{table:compare_2ddb_two_stage}
\end{table}

We further show the superiority of NeuralAnnot in Table~\ref{table:compare_2ddb_two_stage} by comparing it with the two-stage annotators, such as SPIN and EFT.
To this end, we change NeuralAnnot to a two-stage annotator by jointly training it on 3D pseudo-GTs of MoCap datasets (\textit{i.e.}, Human3.6M and MPI-INF-3DHP) and in-the-wild datasets (\textit{i.e.}, MSCOCO), where the 3D pseudo-GTs are obtained from our original one-stage NeuralAnnot.
For the results of SPIN and EFT, we used their officially released 3D pseudo-GTs.
The comparisons show that NeuralAnnot produces 3D pseudo-GTs with the lowest indirect 3D annotation error.
We think this is because our initial 3D pseudo-GTs, obtained by our original one-stage NeuralAnnot, are much beneficial than those of SPIN and EFT, obtained by the optimization-based annotators.

%% file: src/conclusion.tex
\section{Conclusion}

We present NeuralAnnot, the first neural network-based one-stage annotator.
NeuralAnnot is the first attempt to dedicate a neural network for the 3D human model parameter annotation.
It is trained and tested on the same training set with GT 2D/3D joint coordinates in the pre-processing stage of 3D human mesh regressors.
After the pre-processing stage, the outputs of NeuralAnnot on the training set become 3D pseudo-GTs, used to train the regressors.
NeuralAnnot produces much more accurate and beneficial 3D pseudo-GTs than the most widely used one-stage annotator, SMPLify-X.
We will release our 3D pseudo-GTs, which will be highly beneficial to the community.

%% file: src/suppl.tex
\clearpage

\begin{center}
\textbf{\large Supplementary Material for\\``NeuralAnnot: Neural Annotator for 3D Human Mesh Training Sets"}
\end{center}

\setcounter{figure}{0}
\setcounter{table}{0}
\setcounter{section}{0}

\renewcommand{\thefigure}{\Alph{figure}}
\renewcommand{\thetable}{\Alph{table}}
\renewcommand{\thesection}{\Alph{section}}

In this supplementary material, we present more experimental results that could not be included in the main manuscript due to the lack of space.

\section{Qualitative results}
\subsection{3D body-only pseudo-GTs}
\begin{figure*}
\begin{center}
\includegraphics[width=1.0\linewidth]{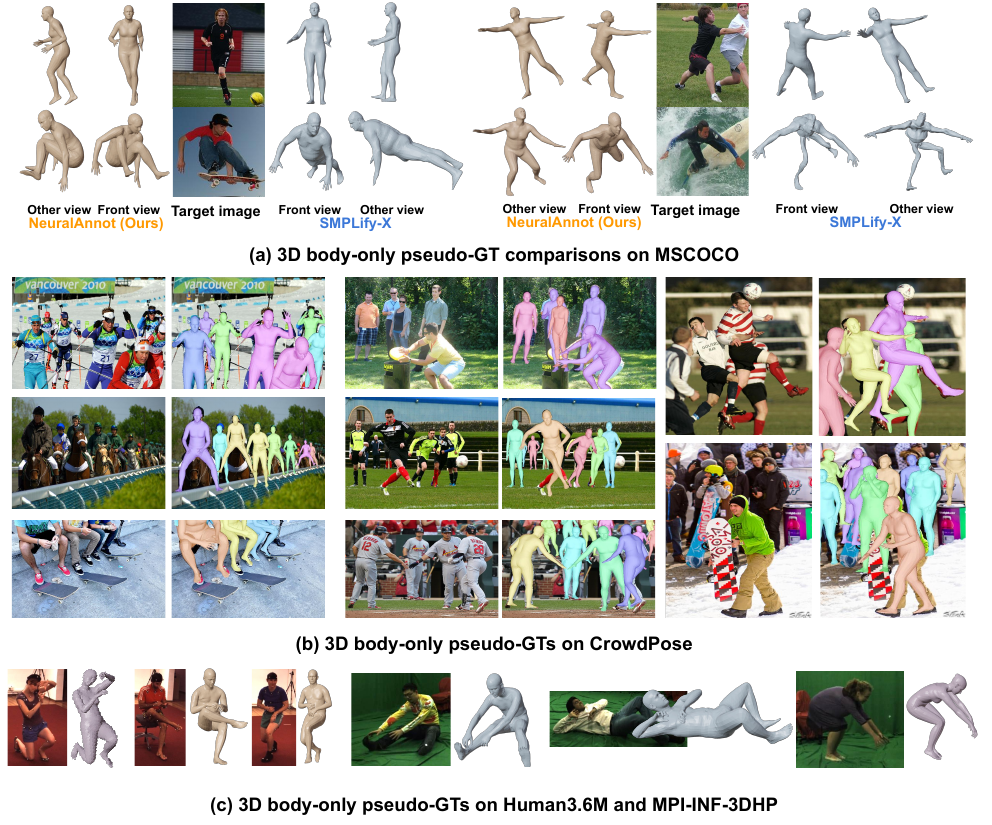}
\end{center}
\vspace*{-3mm}
   \caption{
 (a) Qualitative comparisons between 3D body pseudo-GTs of NeuralAnnot and SMPLify-X on MSCOCO.
 (b) Visualized 3D body pseudo-GTs of NeuralAnnot on CrowdPose.
 (c) Visualized 3D body pseudo-GTs of NeuralAnnot on Human3.6M and MPI-INF-3DHP.
   }
\vspace*{-3mm}
\label{fig:qualitative_body}
\end{figure*}

Figure~\ref{fig:qualitative_body} (a) shows NeuralAnnot produces far better 3D body pseudo-GTs than SMPLify-X.
In particular, it produces much better results when the poses in input images have truncations and complicated articulations.
Figure~\ref{fig:qualitative_body} (b) shows our NeuralAnnot is highly robust to occlusions and truncation in crowd scenes of CrowdPose~\cite{li2019crowdpose}.
Figure~\ref{fig:qualitative_body} (c) shows additional results on Human3.6M~\cite{ionescu2014human3} and MPI-INF-3DHP~\cite{mehta2017monocular}.

\subsection{3D hand-only pseudo-GTs}
\begin{figure*}
\begin{center}
\includegraphics[width=1.0\linewidth]{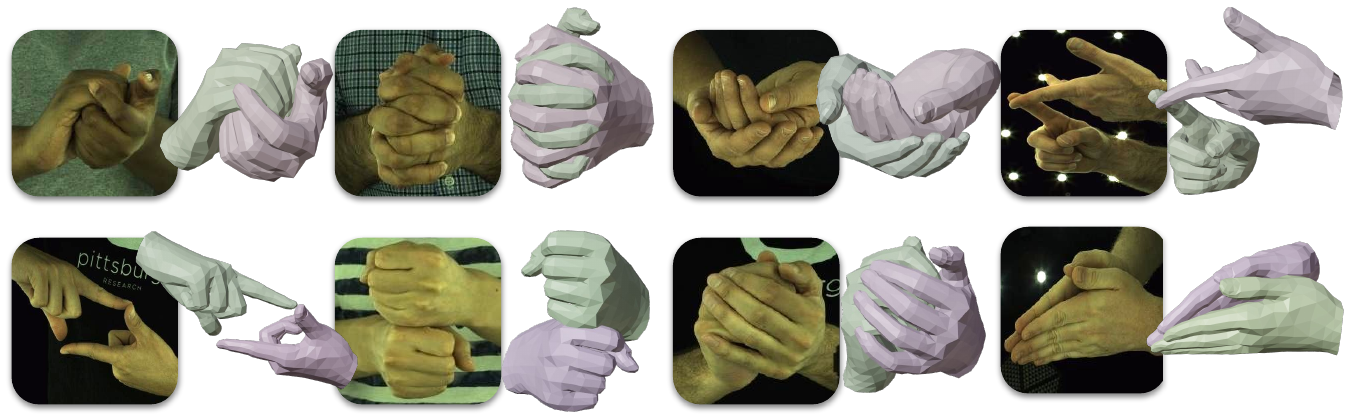}
\end{center}
\vspace*{-3mm}
   \caption{
 Visualized 3D hand pseudo-GTs of NeuralAnnot on InterHand2.6M.
   }
\vspace*{-3mm}
\label{fig:qualitative_hand}
\end{figure*}

Figure~\ref{fig:qualitative_hand} shows 3D hand pseudo-GTs of our NeuralAnnot on InterHand2.6M~\cite{moon2020interhand2}.
It successfully produces 3D pseudo-GTs from highly complicated interacting hand images.

\subsection{3D face-only pseudo-GTs}
\begin{figure*}
\begin{center}
\includegraphics[width=1.0\linewidth]{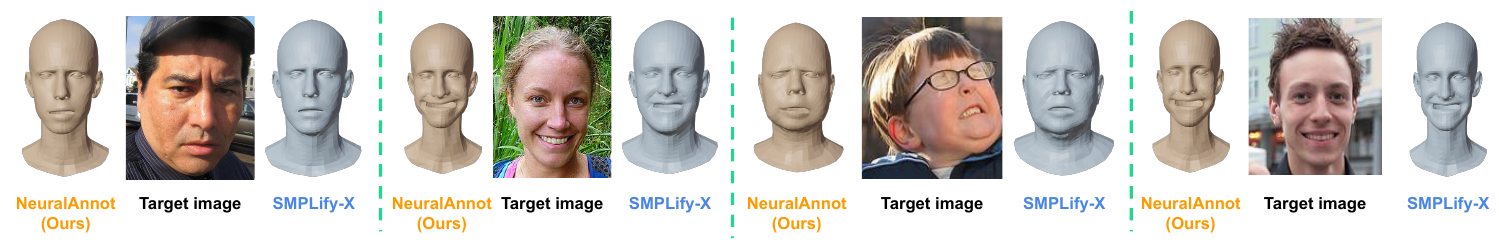}
\end{center}
\vspace*{-3mm}
   \caption{
 Qualitative comparisons between 3D face pseudo-GTs of NeuralAnnot and SMPLify-X on MSCOCO.
 We normalized the global rotation of the face for visualization purpose.
   }
\vspace*{-3mm}
\label{fig:qualitative_face}
\end{figure*}

Figure~\ref{fig:qualitative_face} shows that NeuralAnnot and SMPLify-X produce similar 3D face pseudo-GTs.
Unlike the body and hand parts, the face part does not involve complicated articulation, which makes SMPLify-X work well and produce similar results to those of NeuralAnnot.

\subsection{3D whole-body pseudo-GTs}
\begin{figure*}
\begin{center}
\includegraphics[width=1.0\linewidth]{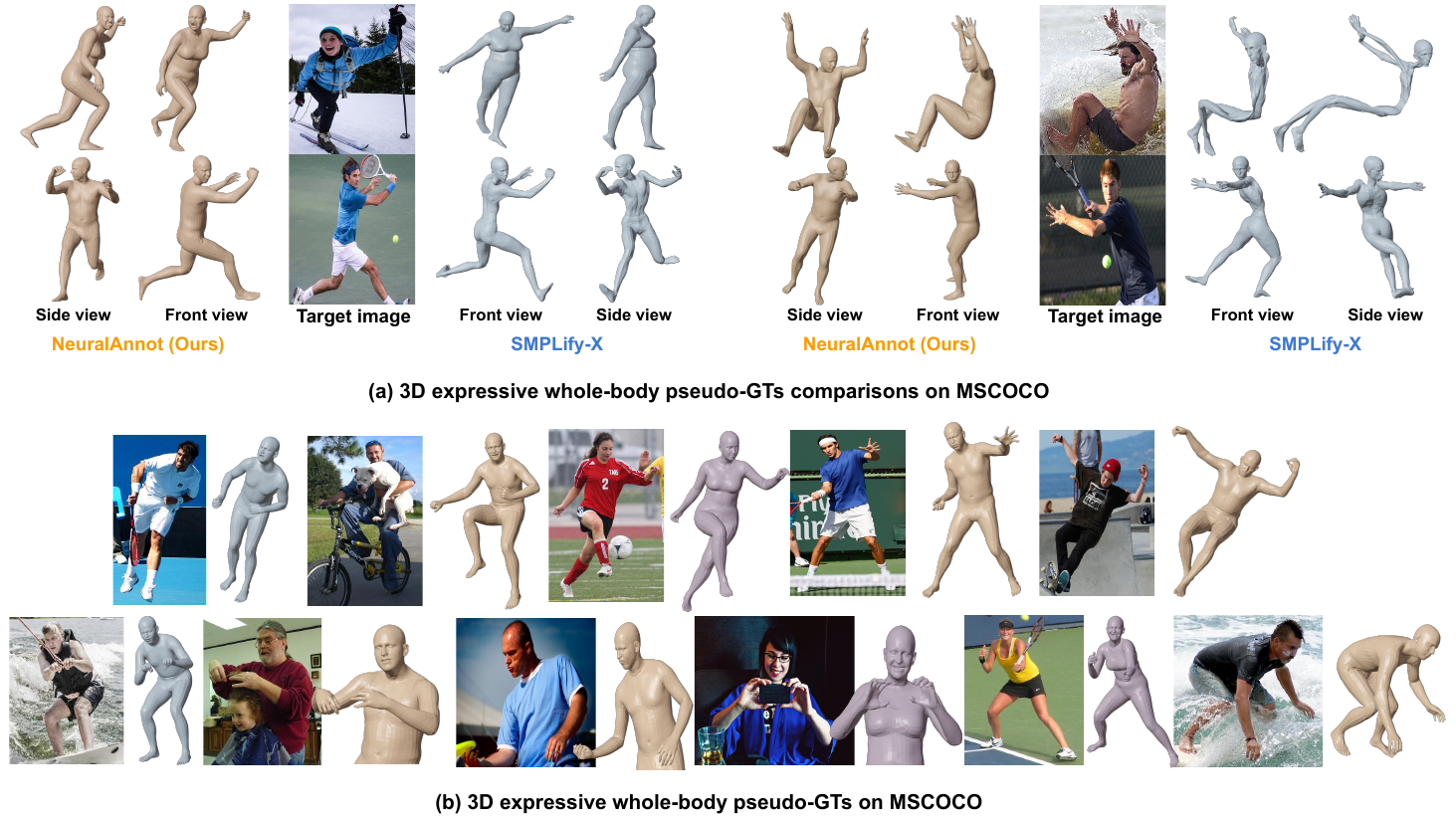}
\end{center}
\vspace*{-3mm}
   \caption{
 Visualized expressive whole-body 3D pseudo-GTs of NeuralAnnot on MSCOCO.
   }
\vspace*{-3mm}
\label{fig:qualitative_expressive}
\end{figure*}

Figure~\ref{fig:qualitative_expressive} (a) shows NeuralAnnot produces much better expressive whole-body 3D pseudo-GTs than SMPLify-X on MSCOCO.
Figure~\ref{fig:qualitative_expressive} (b) shows more qualitative results of NeuralAnnot on MSCOCO.
%We believe the reason why NeuralAnnot produces better 3D pseudo-GTs than SMPLify-X is that the data-driven learning enables NeuralAnnot's network to extract deep semantic features, which provide highly useful cues to produce 3D pseudo-GTs.
%On the other hand, SMPLify-X fits 3D human model parameters to GT of each sample independently without learning useful knowledge from a large number of samples.

\section{Running SMPLify-X on 3D joint coordinates}
As original SMPLify-X does not consider 3D joint coordinates during the optimization, we modified it to consider 3D joint coordinates for 3D pseudo-GTs of Human3.6M~\cite{ionescu2014human3}, MPI-INF-3DHP~\cite{mehta2017monocular}, and InterHand2.6M~\cite{moon2020interhand2}, which provide GT 3D joint coordinates.
To this end, we made two modifications.

\noindent\textbf{Camera initialization.}
We initialize extrinsic camera parameters $R$ and $t$ using hip and shoulder 3D joint coordinates by performing SVD.
For the 3D pseudo-GTs of hands, we use five hand joints, which include the wrist, index root, middle root, ring root, and pinky root.
$R$ and $t$ represent a 3D rotation matrix and 3D translation vector, respectively, from a human model coordinate system to a dataset coordinate system.
We chose the hip and shoulder joints or the five hand joints as they can roughly decide the 3D global rotations of the human body or hands, respectively, while end-point joints (\textit{e.g.}, wrists and ankles for the body and fingertips for the hands) cannot.

\noindent\textbf{3D data term.}
We changed the 2D data term of SMPLify-X to the 3D data term, which calculates a distance between the GT 3D joint coordinates and 3D joint coordinates from a mesh.
The 3D joint coordinates from a mesh are obtained by a joint regression matrix, defined in human models.
We use a Geman-McClure error function~\cite{geman1987statistical} for the distance used in the 2D data term of the original SMPLify-X.
We tried several other distances, such as $L1$ and $L2$, and found that Geman-McClure error function~\cite{geman1987statistical} and $L1$ work the best.
The distance is calculated in a meter scale, and we set the weight of the data term to $10^6$, which works the best.

\section{Qualitative comparisons between two-stage NeuralAnnot and EFT}
\begin{figure*}[t]
\begin{center}
\includegraphics[width=1.0\linewidth]{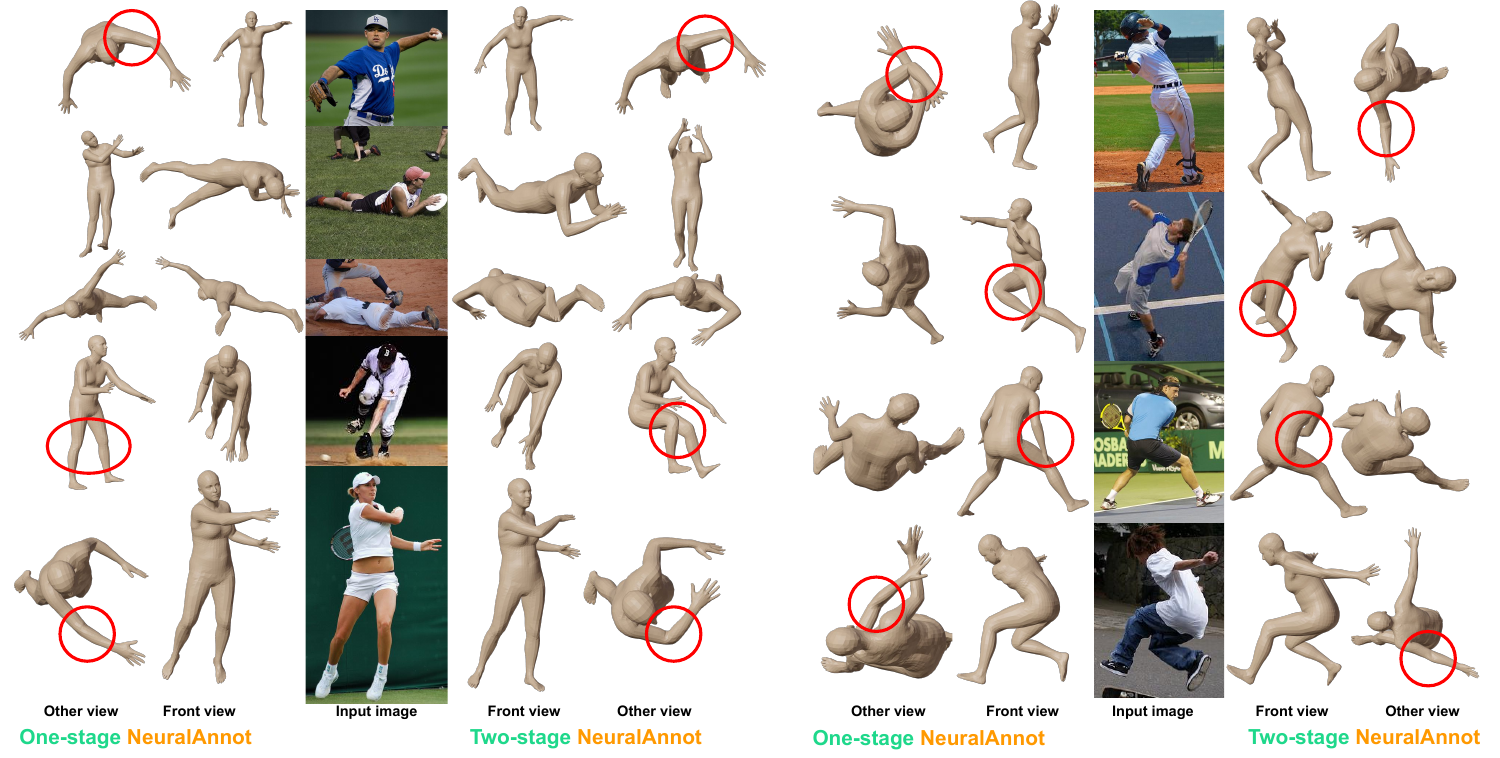}
\end{center}
\vspace*{-3mm}
   \caption{
 Qualitative comparison between one-stage NeuralAnnot and two-stage NeuralAnnot on MSCOCO.
   }
\vspace*{-3mm}
\label{fig:two_stage_neuralannot}
\end{figure*}

\begin{figure*}
\begin{center}
\includegraphics[width=1.0\linewidth]{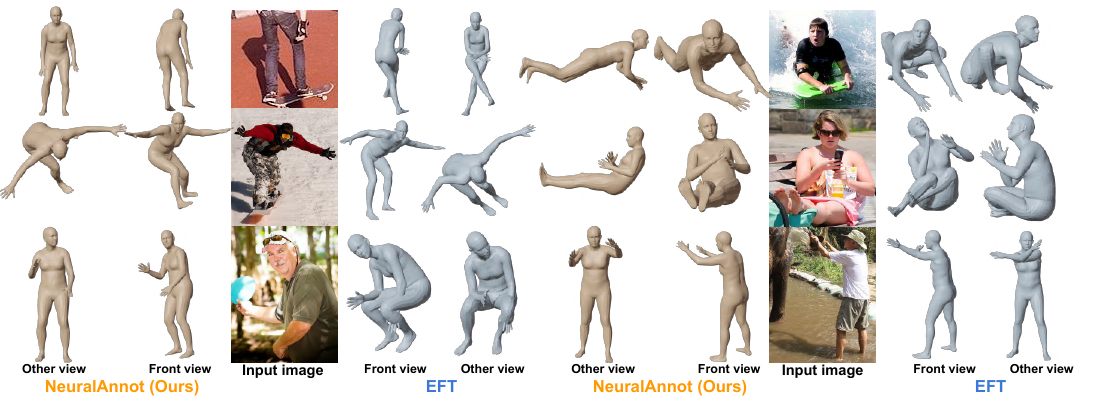}
\end{center}
\vspace*{-3mm}
   \caption{
 Qualitative comparisons between NeuralAnnot and EFT on MSCOCO.
   }
\vspace*{-3mm}
\label{fig:qualitative_comparison_eft}
\end{figure*}

In Table 7 of the main manuscript, we showed our two-stage NeuralAnnot produces more beneficial 3D pseudo-GTs than previous two-stage annotators, such as SPIN~\cite{kolotouros2019learning} and EFT~\cite{joo2020exemplar}.
To this end, we changed our NeuralAnnot to a two-stage annotator by training it on initial 3D pseudo-GTs of Human3.6M, MPI-INF-3DHP, and MSCOCO, where the initial 3D pseudo-GTs are obtained by our original one-stage NeuralAnnot.
Figure~\ref{fig:two_stage_neuralannot} shows that changing our original one-stage NeuralAnnot to a two-stage annotator can correct possibly wrong 3D pseudo-GTs to better ones.
Figure~\ref{fig:qualitative_comparison_eft} shows that our two-stage NeuralAnnot produces better 3D pseudo-GTs than EFT.

We believe there are two reasons why NeuralAnnot produces better ones.
First, EFT is based on a pre-trained SPIN network, trained on 3D pseudo-GTs of SMPLify-X.
SMPLify-X often suffers from inaccurate 3D pseudo-GTs, as shown in Figure 4 of the main manuscript and Figure~\ref{fig:qualitative_body} (a), which can affect 3D pseudo-GTs of SPIN and EFT.
Second, as EFT fine-tunes a pre-trained SPIN network to 2D joint coordinates of each sample, it might produce inaccurate 3D pseudo-GTs when the input image has truncated or invisible joints.
During the fine-tuning, their network can be overfitted to partial joints of a sample as truncated or invisible joints do not have 2D joint coordinates.
Hence, the pre-trained SPIN network can be corrupted and produce wrong 3D poses for truncated or invisible joints as no supervisions are applied for those joints.
On the other hand, our NeuralAnnot's network is not optimized for a specific sample; instead, it is optimized for entire samples of datasets.
Therefore, it does not suffer from the overfitting to a specific sample and produces robust 3D pseudo-GTs when input images have truncated or invisible joints.
The effects of the truncated or invisible joints are shown in the 1) first row and first column, 2) first row and second column, and 3) third row and first column of the Figure~\ref{fig:qualitative_comparison_eft}.
Figure~\ref{fig:qualitative_body} (b) additionally shows that NeuralAnnot produces robust 3D pseudo-GTs under severe truncations.

\subsection*{License of the Used Assets}

\begin{compactitem}[$\bullet$]
    \item MSCOCO dataset~\cite{lin2014microsoft} belongs to the COCO Consortium and are licensed under a Creative Commons Attribution 4.0 License.
    \item InterHand2.6M dataset~\cite{moon2020interhand2} is CC-BY-NC 4.0 licensed.
    \item Human3.6M dataset~\cite{ionescu2014human3}'s licenses are limited to academic use only. 
    \item MPI-INF-3DHP dataset~\cite{mehta2017monocular} is released for academic research only and it is free to researchers from educational or research institutes for non-commercial purposes.
    \item 3DPW dataset~\cite{von2018recovering} is released for academic research only and it is free to researchers from educational or research institutes for non-commercial purposes.
    \item FreiHAND dataset~\cite{Freihand2019} is released for academic research only and it is free to researchers from educational or research institutes for non-commercial purposes.
    \item CrowdPose dataset~\cite{li2019crowdpose} is released for academic research only and it is free to researchers from educational or research institutes for non-commercial purposes.
    \item \href{https://github.com/vchoutas/smplify-x}{SMPLify-X~\cite{pavlakos2019expressive} codes} are released for academic research only and it is free to researchers from educational or research institutes for non-commercial purposes.
    \item \href{https://github.com/facebookresearch/eft}{EFT~\cite{joo2020exemplar} codes} are CC-BY-NC 4.0 licensed.
    \item \href{https://github.com/nkolot/SPIN}{SPIN~\cite{kolotouros2019learning} codes} are released for academic research only and it is free to researchers from educational or research institutes for non-commercial purposes.
\end{compactitem}

